\pdfoutput=1

\documentclass[11pt,a4paper]{article}

\usepackage{times,latexsym}
\usepackage{url}
\usepackage[T1]{fontenc}

\usepackage[acceptedWithA]{tacl2021v1}

\usepackage{xspace,mfirstuc,tabulary}

\newif\iftaclinstructions
\taclinstructionsfalse 
\iftaclinstructions
\renewcommand{\confidential}{}
\renewcommand{\anonsubtext}{(No author info supplied here, for consistency with
TACL-submission anonymization requirements)}
\newcommand{\instr}
\fi

\iftaclpubformat 

\else

\fi

\usepackage{graphicx}
\usepackage{multirow}
\usepackage{booktabs}
\usepackage{amsmath}
\usepackage{amssymb}
\usepackage{array}
\usepackage{float}
\usepackage[font=small,skip=6pt]{caption}
\usepackage{microtype}

\newcommand{\datasetname}{\mbox{\textsc{AmbiFC}}}

\newcommand{\datacertain}{\datasetname{}$^C$}

\newcommand{\datauncertain}{\mbox{\datasetname{}$^{U}$}}

\title{\datasetname{}: Fact-Checking Ambiguous Claims with Evidence}

\author{
 Max Glockner$^{1,5}$,
 Ieva Staliūnaitė$^2$,
 James Thorne$^3$,
 Gisela Vallejo$^4$, \\
 \textbf{Andreas Vlachos}$^2$\ and
 \textbf{Iryna Gurevych}$^{1,5}$ \vspace{0.25em}\\
  $^1$UKP Lab,  Department of Computer Science, Technical University of Darmstadt, \\
  $^2$Department of Computer Science and Technology, University of Cambridge,
  \\
  $^3$KAIST AI, $^4$The University of Melbourne, $^5$hessian.ai
  \\
  \texttt{\{max.glockner,iryna.gurevych\}@tu-darmstadt.de}, \\
  \texttt{\{irs38,av308\}@cam.ac.uk}, \texttt{thorne@kaist.ac.kr},\\
  \texttt{gvallejo@student.unimelb.edu.au} 
}

\date{}

\begin{document}
\maketitle

\begin{abstract}

Automated fact-checking systems verify claims against evidence to predict their veracity. In real-world scenarios, the retrieved evidence may not 
unambiguously support or refute
the claim and yield 
conflicting but valid
interpretations. Existing fact-checking datasets assume that the models developed with them 
predict a single veracity label for each claim, 
thus discouraging the handling of
such ambiguity. To address this issue we present \datasetname{}\footnote{\url{https://github.com/CambridgeNLIP/verification-real-world-info-needs}}, a 
fact-checking dataset with 10k claims derived from real-world information needs. It contains fine-grained evidence annotations of 50k passages from 5k 
Wikipedia pages. We 
analyze the disagreements arising from 
ambiguity when comparing claims against evidence 
in \datasetname{}, observing a strong correlation of annotator disagreement 
with
linguistic phenomena such as underspecification and probabilistic reasoning. 
We develop models for predicting veracity handling this ambiguity
via soft labels, and
find that a pipeline 
that learns the label distribution
for sentence-level evidence selection and veracity prediction yields the best performance. 
We compare models trained on different subsets of \datasetname{} and show that models trained on the ambiguous instances perform better when faced with 
the identified linguistic phenomena. 

\end{abstract}

\section{Introduction}
In Natural Language Processing, the task of automated fact-checking is given a claim of unknown veracity, to identify evidence from a corpus of documents, and predict whether the evidence supports or refutes the claim.
It has received considerable attention in recent years \citep{guo-etal-2022-survey} and gained renewed relevance due to the hallucination of unsupported or even false statements in natural language generation tasks, including information-seeking dialogues \citep{dziri2022faithdial, ji2023survey}. 

Automated fact-checking is closely related to natural language inference (NLI) where the evidence is considered given \citep{thorne-etal-2018-fever, wadden-etal-2020-fact, schuster-etal-2021-get}. 
Several studies \citep{pavlick-kwiatkowski-2019-inherent,nie-etal-2020-learn,jiang-tacl2022-investigating} have shown that  NLI  suffers from inherent ambiguity leading to conflicting yet valid annotations. 
To address this, recent work has focused on utilizing these conflicting annotations, especially when aggregated labels are not considered to adequately represent the task \citep{plank-2022-emnlp,leonardelli2023semeval2023}.

\begin{figure*}
\small
    \centering
    \includegraphics[width=\textwidth]{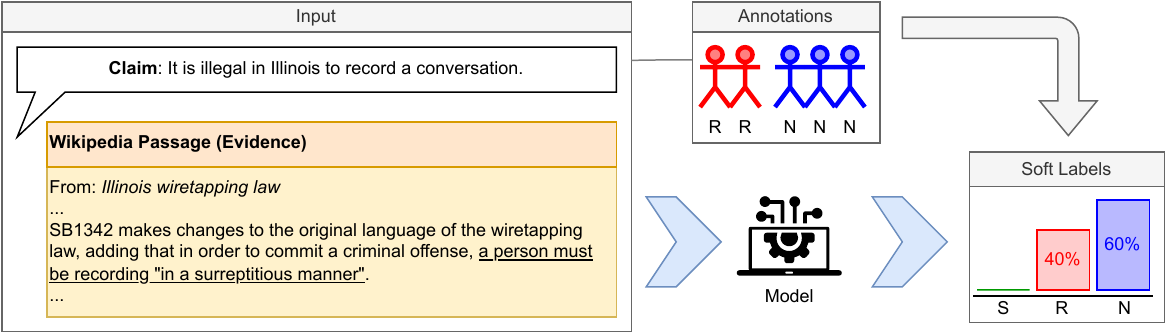}
    \caption{An example of an instance of claim and Wikipedia passage which is ambiguous due to underspecification. We consider all supporting (S), refuting (R) and neutral (N) annotations as valid perspectives. Given a claim and a Wikipedia passage, the model must predict soft labels derived from these annotations.}  
    \label{fig:figure-task-examples}
\end{figure*}
Many fact-checking datasets are purpose-made rather than naturally occurring, similar to those used in NLI; their claims are often created by manipulating 
sentences from the evidence documents \citep{thorne-etal-2018-fever, jiang-etal-2020-hover, Aly21Feverous}. 
As a result, they are unlikely to represent real-world information needs, as they are written with knowledge of the evidence.
On the other hand, in datasets with real-world claims evidence is often used without manual annotation, assuming that it is sufficient \citep{glockner2022missing}. 
If evidence annotation is performed, datasets include artificially created incorrect claims, ensuring that the used evidence contradicts the claims \citep{wadden-etal-2020-fact, saakyan-etal-2021-covid}, or exhibit low annotator agreement \citep{hanselowski-etal-2019-richly, diggelmann2020climate} without attempts to handle ambiguity.
However, even human fact-checkers often disagree, particularly in ambiguous cases \citep{lim2018checking}. 

More concretely,
the claim that \textit{``it is illegal in Illinois to record a conversation''} in Figure~\ref{fig:figure-task-examples} seems clear on its own, yet becomes ambiguous when compared to the evidence, 
as it is underspecified. 
The claim does not explicitly state whether the recording was done surreptitiously (i.e.\ secretively), allowing for various interpretations: 
(a) as refuting the claim since it is legal if not done surreptitiously, and (b) as neutral as it is impossible to determine whether it refutes or supports the claim without information about the recording intent.  
Surreptitious recording only pertains to a specific case and none of the annotators deemed it as prominent enough to provide overall support for the claim.

In this study we aim to investigate the presence of such ambiguities in fact-checking using realistic claims and evidence. 
To this end,
we present \datasetname{}, a large fact-checking dataset derived from real-world information needs, sourced from real-world yes/no questions of BoolQ~\citep{clark-etal-2019-boolq}. \datasetname{} contains evidence annotations at the passage and sentence level from full Wikipedia pages, from
a minimum of five annotations per instance.
Unlike previous fact-checking datasets we consider each annotation as a valid perspective of the claim's veracity given a Wikipedia passage as evidence, and task models to predict the veracity via soft labels that consider all annotations.
We provide explanations for the annotator disagreement via our 
annotations of linguistic phenomena, inspired by \citet{jiang-tacl2022-investigating}, adding inference types idiosyncratic to fact-checking.
Further, we 
experiment with
three established methods to model 
annotator disagreement.  
Our work emphasizes the importance of ambiguity within automated fact-checking and takes a step towards incorporating ambiguity into fact-checking models.

\section{Related Work}
\label{sec:related-work}
Disagreement among humans are often studied in
computational argumentation. 
\citet{habernal-gurevych-2017-argumentation} create a realistic dataset for mining arguments from online discussions, covering various topics. Perspectrum~\citep{chen-etal-2019-seeing} gathers different perspectives supported by evidence and their stance on claims. 
However, computational argumentation focuses on controversial topics with diverse legitimate positions, while automated fact-checking focuses on claim factuality.

In automated fact-checking, earlier works constructed complex claims from question answering datasets \citep{jiang-etal-2020-hover, tan-etal-2023-multi2claim, park2021faviq} or knowledge graphs \citep{kim2023factkg}.
Our work is most comparable to FaVIQ~\citep{park2021faviq}, which was also  generated from real-world information needs questions. Unlike \datasetname{}, it lacks  evidence annotations and utilizes 
disambiguated question-answer pairs from AmbiQA~\citep{min-etal-2020-ambigqa},
hence excluding the natural ambiguity of claims based on real-world information needs, studied in this work.

 Other works gathered claims from credible sources such as scientific publications or Wikipedia, using cited documents as evidence. This  provides  realistic claims which are only supported by evidence, and requires the generation of artificial refuted claims \citep{sathe-etal-2020-automated,wadden-etal-2020-fact,saakyan-etal-2021-covid}, or only distinguishes between different levels of support \citep{kamoi2023wice}.
Another line of research collects claims from professional fact-checking organizations. These works often face disagreement among annotators but do not handle ambiguity \citep{hanselowski-etal-2019-richly,sarrouti-etal-2021-evidence-based}, or do not provide annotated evidence \citep{augenstein-etal-2019-multifc,khan-etal-2022-watclaimcheck}. 
The recently published AVeriTeC dataset \citep{schlichtkrull2023averitec} reconstructs the fact-checkers' reasoning via questions and answers from evidence documents. 
In \datasetname{} we consider claims  that are interesting according to the search queries used in constructing BoolQ \citep{clark-etal-2019-boolq}, not 
claims deemed check-worthy by fact-checkers.
Additionally, we provide passage- and sentence-level annotation, and address uncertainty and disagreement.

In the domain of NLI, \citet[ChaosNLI]{nie-etal-2020-learn} presented a comprehensive annotation of NLI items, involving 100 annotators for each item.
\citet{jiang-tacl2022-investigating} further investigate the causes of disagreement in ChaosNLI, categorizing them into pragmatic and lexical features, as well as general patterns of human behavior under annotation instructions. 
Our work extends the existing work in NLI to fact-checking, by examining the types of linguistic phenomena common in the two tasks.
\citet{plank-2022-emnlp} and \citet{uma2021learning} provide overviews of the current state of modeling and evaluation techniques for data with annotation variance. They highlight various methods, such as calibration, sequential fine-tuning, repeated labeling, learning from soft labels, and variants of multi-task learning. 

\section{Preliminaries}

 Each instance  $(c, P)$  comprises a claim $c$ and a passage $P$ from Wikipedia. A passage $P = [s_1, s_2, ..., s_n]$ is composed of $n$ sentences $s_i$.
 Annotations are collected for the entire passage $P$ and for each individual $s_i \in P$, indicating their stance towards $c$ as \emph{supporting}, \emph{refuting} or \emph{neutral}.
These \emph{ternary} sentence-level annotations expressing stance towards the claim can be mapped to \emph{binary} annotations by treating non-neutral annotations as ``evidence'' regardless of stance.
We do not aggregate passage-level annotations into hard veracity labels. Instead, for each ($c$, $P$) we use soft labels, representing the veracity as a distribution of the passage-level annotations given a claim.

We specifically focus on the fact-checking subtasks of Evidence Selection (Ev.) and Veracity Prediction (Ver.) for each claim-passage instance ($c$ $P$). 
We consider each sentence $s_i$ as part of the evidence $E$ for $c$ if at least one non-neutral annotation for it exists.
For the evidence selection subtask, the model must select all evidence sentences $s_i \in E$ in $P$. 
In the veracity prediction subtask, the model must predict the veracity of $c$ given $P$ using soft labels that represent the annotation distribution at the passage level (Figure~\ref{fig:figure-task-examples}). In addition to comparing the predicted and human label distributions, we assess the models using less stringent metrics (outlined in §\ref{sec:experiments:veracity-prediction-soft-labels}) to accommodate potential annotation noise.

\section{The \datasetname{} dataset}
\label{sec:data-collection}

To create the claims and annotate them with evidence, we followed a two-step process. First, crowd-workers transformed questions from BoolQ into assertive statements. Second, the crowd-workers labeled evidence sentences and passages from a Wikipedia page to indicate whether they support or refute the corresponding claim.

\paragraph{Claims}
BoolQ comprises knowledge-seeking user queries with yes-or-no answers, similar to fact-checking intentions. 
Dataset instances are generated by rephrasing these queries into claims. 
Two annotators on Mechanical Turk rephrase each BoolQ question as a claim, with instructions to retain as many tokens from the original question as possible. 
In case the claims by the two annotators were different, they were
 included in the dataset after manual review. The crowd-workers underwent a qualification round evaluated by the authors. 512 unique annotators with a 95\% acceptance rate completed the task. 20\% of HITs were used for worker qualification and training, 80\% form the final dataset.

\paragraph{Evidence Annotation}
\label{sec:evidence}
For each claim, the full Wikipedia page from BoolQ containing the answer to the yes/no question, was used as evidence. To prevent positional bias, where annotators concentrate on a page's beginning,
and annotator fatigue, pages were divided into multiple passage-level annotation tasks (capped at 20 contiguous sentences). 
Annotators assessed each sentence
in a passage as \emph{supporting}, \emph{refuting} or \emph{neutral} towards the claim, and 
provided an overall judgment of the claim's veracity given the passage. Passages without evidence sentences were labeled neutral.
In anticipation of potentially low inter-annotator agreement as observed in comparable annotation tasks \citep{hanselowski-etal-2019-richly, diggelmann2020climate}, we introduce a second level of passage annotation to indicate uncertainty:
If annotators chose ``neutral'' they could additionally flag passages as ``relevant'' to differentiate it from entirely unrelated passages.
Non-neutral passage annotations could be flagged as ``uncertain'' by the annotators.
We treat both of these additional labels (``relevant'' for ``neutral'' instances and ``uncertain'' for non-neutral ones)
as indicators of unclear decision boundaries.
Passages received two initial annotations, with an additional three 
for passages with at least one supporting or refuting initial annotation, resulting in five annotations per instance in these cases. Instances with identical claims (from identical paraphrasing of questions by different annotators) and passages were merged, resulting in instances with more than five annotations. 

\paragraph{Quality Controls}
Annotators underwent a 3-stage approval process consisting of a qualification quiz, manual review of their first 100 HITs and continuous manual review. Errors were communicated to them 
to provide formative feedback. 
A batch of claims was sampled daily for continuous manual review during annotation. The authors reviewed and accepted 12,137 HITs (5.2\% of all annotation tasks), while corrections were provided for additional 400 HITs, indicating a 3.2\% error rate where annotators deviated from guidelines, \emph{not} due to differences in opinion. The number of HITs reviewed for each annotator was proportional to the annotator's error rate and the number of annotations submitted. 
Annotation times were used to calibrate worker hourly pay at \$22.

\paragraph{Agreement}
The inter-annotator agreement in terms of Krippendorff's $\alpha$ on the collected data is 0.488 on the passage veracity labels and 0.394 on the sentence level. The disagreement implies that single labels cannot capture all valid viewpoints, necessitating the use of soft labels 
for evaluation.
Fully neutral samples have only two annotations (as per dataset construction), which is insufficient for reliable evaluation of soft labels, unless we can ensure that they are indeed 100\% neutral. 
We estimate the probability of misclassifying an instance as fully neutral when only seeing two annotations, by randomly selecting two annotations from samples with 5+ annotations. 
The likelihood of wrongly assuming an instance as fully neutral when observing two neutral annotations is 0.9\% for the entire dataset 
but it increases up to 20.9\% when sampling from uncertain instances.
Using this estimate, we omit fully neutral (but ``relevant'') instances from our experiments, while retaining them in our linguistic analysis in §\ref{sec:analaysis:disagreement-analysis}.

\paragraph{Subsets of \datasetname{}}
\begin{table}[t]
\small
    \centering
    \begin{tabular}{l | c c | c}

    & \multicolumn{2}{c|}{\textbf{\datacertain{}}} & \multicolumn{1}{c}{\textbf{\datauncertain{}}} \\
    & \textit{2-4 Ann.} & \textit{5+ Ann.}  & \textit{5+ Ann.} \\
    \toprule
    \textbf{Claims} & 6,241 & 4,613 & 9,380 \\
    \textbf{Wiki Pages} & 3,418 & 2,732 & 4,789 \\
    \textbf{Cl./Passage} & 18,214 & 6,475 & 26,680 \\
    \textbf{Cl./Sentence} & 141,079 & 49,497 & 223,370 \\
    \midrule
    \textbf{Pass. Ann.} &&&\\
    \textit{Has N} &100 \%& 38.7 \%&93.1 \% \\
    \textit{Has S} &0 \%& 82.2 \%& 78.4 \%\\
    \textit{Has R} &0 \%& 29.0\%&  42.0 \%\\
    \textit{Has S} \& \textit{R} &0 \%& 11.3 \%&  21.6 \%\\
      \bottomrule
    \end{tabular}
    \caption{\datasetname{} statistics including passages containing \textbf{S}upporting, \textbf{R}efuting and/or \textbf{N}eutral annotations. }
    \label{tbl:experiments-data-statistics-new}
\end{table}
We partitioned 
instances into subsets based on the additional labels "relevant" (for neutral passages) and "uncertain" (for non-neutral passages) provided by the annotators. 
Instances marked with either of these labels by any annotator form the ``uncertain'' subset (\datauncertain{}), while the remaining instances form the ``certain'' subset (\datacertain{}).
We split \datasetname{} into train/dev/test splits with the proportions of 70/10/20 for both \datacertain{} and \datauncertain{} based on instances ($c$, $P$). We ensure each Wikipedia page only exists in one split, and that the claims and Wikipedia pages occur in the same split regardless of their belonging to \datacertain{} or \datauncertain{}.
The entire \datasetname{} includes 51,369 instances ($c$, $P$) with 10,722 unique claims and 5,210 unique Wikipedia pages (Table~\ref{tbl:experiments-data-statistics-new}).
Similar to VitaminC~\citep{schuster-etal-2021-get}, each claim is annotated based on different evidence passages. Consequently, the same claim may have differing veracity labels depending on the passages. This helps diminish the influence of claim-only biases \citet{schuster-etal-2019-towards}, and allows the same claim to be present in both subsets with different evidence passages.
For 7,054 claims (65.8\%), ($c$, $P$) instances exist in both subsets, \datauncertain{} and \datacertain{}.
Passages with contradictory veracity labels are substantially more frequent in \datauncertain{} than in \datacertain{} (21.6\% vs 11.3\%).
Instances in \datauncertain{} have at least one non-neutral annotation, as per the dataset annotation process. However, 93.1\% of them additionally contain at least one neutral annotation, indicating possibly insufficient evidence. 
Thus, models cannot achieve high performances when relying on spurious correlations within the claim only \citep{schuster-etal-2019-towards, hansen-etal-2021-automatic}.

\paragraph{Positional Analysis}
\label{sec:positional-analysis}
\begin{figure}
\small
    \centering
    \includegraphics[width=\linewidth]{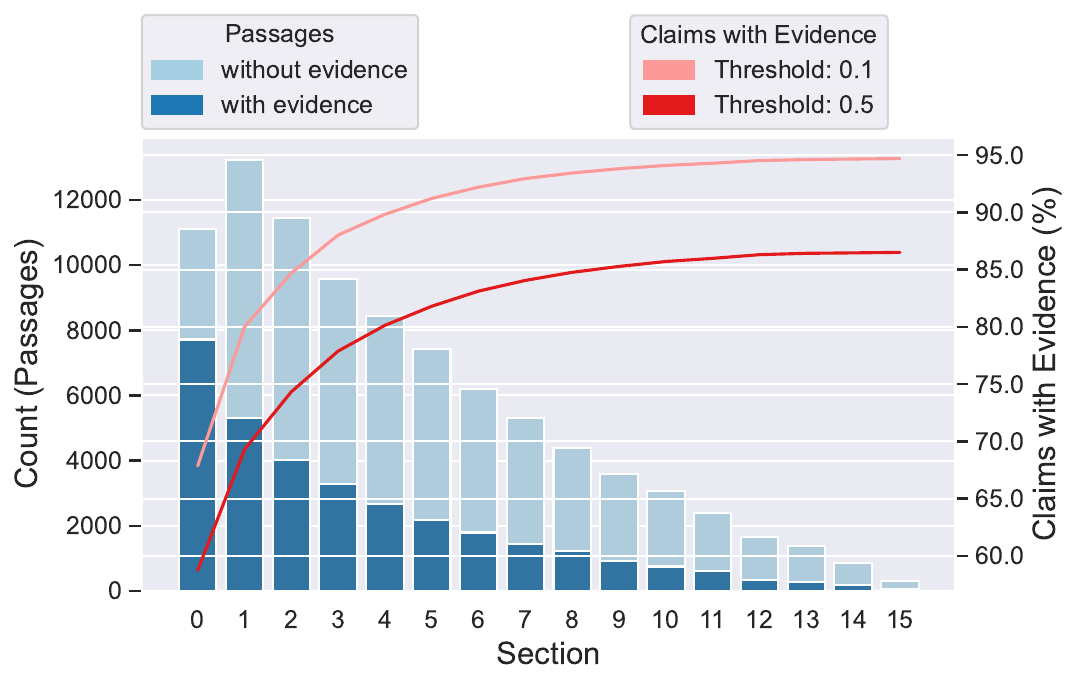}
    \caption{Evidence by section for claims and passages.}
    \label{fig:evidence-by-section}
\end{figure}
Wikipedia pages have general information in the introduction and more specific details in later sections. In contrast to FEVER, which only uses introductions, our approach involves utilizing passages from entire Wikipedia pages.
Figure~\ref{fig:evidence-by-section} visualizes the detected evidence per Wikipedia section, revealing that a substantial number of passages from later sections contain evidence for or against the claim. 
The curves  show cumulatively the number of claims with evidence found per section when considering passages with at least 10\% or 50\% of non-neutral annotations as evidence. While most claims have sufficient evidence in the early sections, there are still many claims that require later sections to be verified.

\section{Disagreement Analysis}
\label{sec:analysis}
\subsection{Quantitative Analysis} 
\label{subsec:agreement}
\begin{table}[t]
\small
    \centering
    \begin{tabular}{c r | c c}
    && \multicolumn{2}{c}{\textit{Krippendorff's} $\alpha$} \\
    \textbf{Label} & \textbf{Samples} & \textbf{\datacertain{}} & \textbf{\datauncertain{}}\\
    \toprule
    \multicolumn{2}{c|}{\textbf{Sentence}} \\
    \textit{binary} & all & 0.607  & ---  \\
    \textit{binary} & 5+ Ann. & 0.563 & 0.314  \\
    \textit{ternary} & all & 0.595  & ---  \\
    \textit{ternary} & 5+ Ann. & 0.560 & 0.302  \\
    \midrule
    \multicolumn{2}{c|}{\textbf{Passage}} \\
    \textit{stance} & all & \textbf{0.815}  & --- \\
    \textit{stance} & 5+ Ann. & 0.553 & \textbf{0.206}  \\
      \bottomrule
    \end{tabular}
    \caption{Krippendorff's $\alpha$ over different subsets.
    Samples in the \textbf{bold} are used for \datasetname{}.}
    \label{tbl:agreement-evidence}
\end{table}

\paragraph{Agreement over Subsets}
Table \ref{tbl:agreement-evidence} shows the agreement results for both subsets. We compared ternary and binary evidence labels at the sentence level. 
The inter-annotator agreement for the  instances (``all'') in \datacertain{} is 0.607 when calculated with binary labels and 0.595 with ternary labels. For the utilized instances in \datauncertain{}, the agreement is 0.314 with binary labels and 0.302 with ternary labels. The minor differences in agreement under both labeling schemes suggest that annotators with conflicting interpretations may  emphasize different evidence sentences rather than assigning opposing labels to the same sentences.
The agreement is consistently higher for the certain subset compared to the uncertain subset. 
The passage-level agreement for the certain subset, measured by Krippendorff's $\alpha$, is 0.815. When computed over instances with 5+ annotations (removing neutral instances with perfect agreement as they did not receive annotations beyond the first two) the agreement drops to 0.553.
We observe much poorer agreement (0.206) on \datauncertain{}.
The difference in inter-annotator agreement between these two subsets, based on the annotators' own judgment, signals their awareness of alternative interpretations on these instances.

\paragraph{Agreement over Sections}
\begin{figure}
\small
    \centering
    \includegraphics[width=\linewidth]{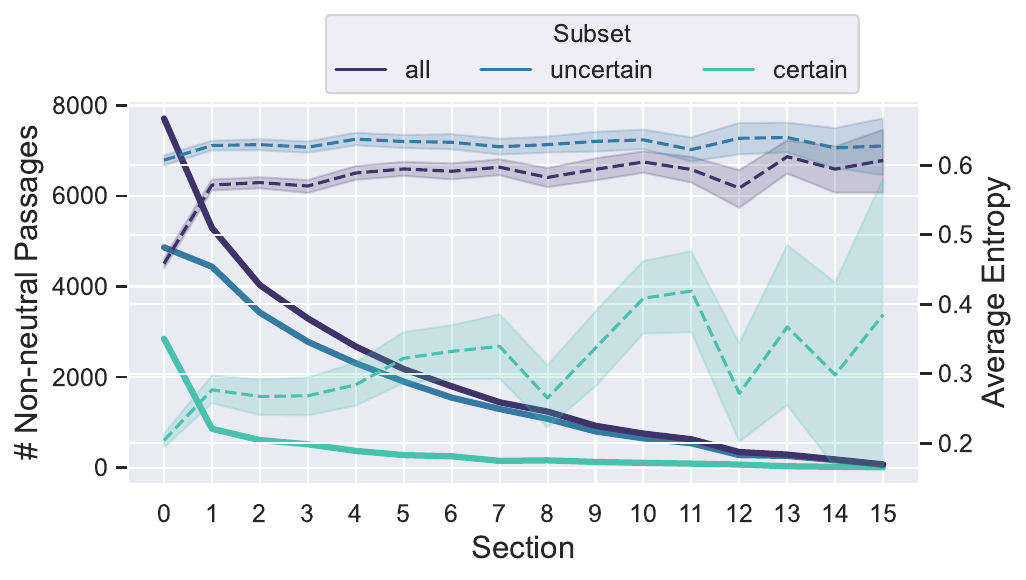}
    \caption{Passage-level annotation entropy (dashed) and count (solid) per section over samples with 5+ annotations.}
    \label{fig:entropy-by-section}
\end{figure}
\label{par:agreementoversections}
Continuing from the positional analysis (§\ref{sec:positional-analysis}), we explore whether the position of evidence passages within sections affects annotator disagreement.
Figure~\ref{fig:entropy-by-section} visualizes the number of instances (solid) and average passage-level annotation entropy (dashed), separated by subset. We only consider passages with 5+ annotations.
The entropy is relatively stable within each subset, 
but substantially different between them.
Instances from \datacertain{} mostly contain evidence in the first section, with few samples in later sections considered certain by annotators. In contrast, instances from \datauncertain{} appear throughout most sections. 

\paragraph{Agreement per Veracity Interpretation}
\label{subsec:analysis:different-evidence}
\begin{table}
    \small
    \centering
    \begin{tabular}{l|ccc}
    \toprule
        \textbf{Subset} & \textbf{Size} &  \textbf{Agreement per}  & $\alpha$ \\
        \midrule

        \multirow{ 2}{*}{\datacertain{} } & \multirow{ 2}{*}{1,000 } &Instance & .404\\
          & & Veracity & .601\\
        \midrule
        \multirow{ 2}{*}{\datauncertain{} } & \multirow{ 2}{*}{14,814 } &Instance & .250\\
         & &Veracity & .561\\
        \midrule
        \multirow{ 2}{*}{\datasetname{} } & \multirow{ 2}{*}{15,814 } & Instance & \textbf{.264}\\
         & & Veracity & \textbf{.565}\\
        
        \bottomrule
    \end{tabular}
    \caption{Sentence-level Krippendorff's $\alpha$ of evidence annotations between all annotators of the same \textbf{Instance}, or \textbf{Veracity} interpretation on the same instance.}
    \label{tbl:agreement-evidence-by-passage-annotation}
\end{table}


        
We aim to determine if different annotators focus on different, or on the same sentences of a passage when assigning contradictory veracity labels to a claim.\footnote{Contradictory sentence- and passage-level annotations by the same annotator  occur only in 0.8\% (389 instances).} 
To examine this, we calculate the agreement among sentence annotations 
over binary evidence labels
in two scenarios:
(1) between all annotations of the same instance ($c$, $P$), and (2) between all annotations of the same instance when annotators assigned the same veracity label $y_p$ to the claim ($c$, $P$, $y_p$). The results are reported in Table~\ref{tbl:agreement-evidence-by-passage-annotation}.
To compare, we need at least two annotators per instance and veracity label. 
This yields annotations for 15,814 ($c$, $P$) instances (32.3\% of all instances with 5+ annotations). Due to this selection, this subset represents a highly ambiguous subset of \datasetname{}. 
As expected, the evidence inter-annotator agreement computed at the instance level is poor.
When only comparing the evidence annotations among annotators who assigned the same veracity label, the agreement is substantially higher. This suggests that annotators deemed different sentences as important when assigning different veracity labels.

\subsection{Linguistic Analysis}
\label{sec:analaysis:disagreement-analysis}

\begin{table*}[htb]
  \centering
\begin{center}
\footnotesize{
\begin{tabular}{ p{8.0cm}p{5.6cm}p{1.5cm}}
 \hline
{\bf Example}&{\bf Inference interpretation} & {\bf Annotations}\\
 \hline   
  \multicolumn{3}{c}{\sc{Implicature}}\\
 \underline{Claim:}  \textit{Red eared sliders can live in the ocean.} \newline \underline{Evidence:} \textit{In the wild, red-eared sliders brumate over the winter at the bottoms of ponds or shallow lakes.} & Listing the types of bodies of water that red ear sliders brumate in implies that the ocean is not one of them. & [N, N, R, R, R, R]\\
 \hline
  \multicolumn{3}{c}{\sc{Presupposition}}\\
 \underline{Claim:} \textit{The Queen Anne's Revenge was a real ship.} \newline \underline{Evidence:}\textit{On June 21, 2013, the National Geographic Society reported recovery of two cannons from Queen Anne's Revenge.} & The evidence presupposes that Queen Anne's Revenge is an existing ship by stating that parts of it were recovered. & [S, S, S, S, N]
\\
 \hline
  \multicolumn{3}{c}{\sc{Coreference}}\\
\underline{Claim:}  \textit{Steve Carell will appear on the office season 9.} \newline \underline{Evidence:} \textit{This is the second season not to star Steve Carell as lead character Michael Scott, although he returned for a cameo appearance in the series finale.} & Whether the claim is supported or refuted depends if 
`series finale' and `season 9' have the same referent. & [S, S, S, S, S, R, R]\\
 \hline
  \multicolumn{3}{c}{\sc{Vagueness}}\\
 \underline{Claim:} \textit{Gibraltar coins can be used in the UK.} \newline \underline{Evidence:} \textit{Gibraltar's coins are the same weight, size and metal as British coins, although the designs are different, and they are occasionally found in circulation across Britain.} & The veracity judgment depends on the meaning of the word `can' being interpreted as `be able to' or `be legally allowed to'. & [S, N, N, N, N, R, R] \\
 \hline
  \multicolumn{3}{c}{\sc{Probabilistic Enrichment}}\\
 \underline{Claim:}  \textit{It is rare to have 6 wisdom teeth.} \newline \underline{Evidence:} \textit{Most adults have four wisdom teeth, one in each of the four quadrants, but it is possible to have none, fewer, or more, in which case the extras are called supernumerary teeth.} & The fact that most adults have 4 wisdom teeth makes it likely that having 6 is rare. & [S, S, S, N, N]\\
 \hline
   \multicolumn{3}{c}{\sc{Underspecification}}\\
\underline{Claim:}  \textit{You cannot have a skunk as a pet.} \newline \underline{Evidence:} \textit{It is currently legal to keep skunks as pets in Britain without a license.} & The claim is false under specific conditions of location, and underspecified otherwise. & [S, N, R, R, R, R, R]\\
 \hline
 \end{tabular}
 }
 \caption{Examples of claim and relevant evidence which require different types of inference to resolve, and their corresponding veracity annotations at the passage level: \textbf{R}efuted, \textbf{N}eutral and \textbf{S}upported.}
\label{tab:pragmaticexamples}
\end{center}
\end{table*}

To assess the extent to which annotator disagreement in \datasetname{} can be attributed to ambiguity, we conduct a statistical analysis that examines various forms of linguistic inference in the data and their relationship to annotator disagreement on veracity labels. 
We hypothesize that lexical, discourse, and pragmatic inference contributes to disagreements. 
The inference classes considered are \textit{Implicature, Presupposition, Coreference, Vagueness, Probabilistic Enrichment} and \textit{Underspecification}, and examples of each type are shown in Table~\ref{tab:pragmaticexamples}.
\textit{Implicature} concerns content that is suggested by means of conversational maxims and convention, but not explicitly stated \citep{gricelogic}.
A \textit{Presupposition} refers to accepted beliefs within a discourse \citep{karttunen1974presupposition}.
\textit{Coreference} is used here as a shorthand for difficulty in resolving coreference of ambiguous denotations \citep{hobbs1979coherence},
\textit{Vagueness} describes terms with fuzzy boundaries \citep{kenney1996vagueness}, and
\textit{Probabilistic Enrichment} is a class for inferences about what is highly likely but not entailed. These classes closely follow the framework of \citet{jiang-tacl2022-investigating}, with changes as follows.

Experimental research has explored the issue of \textit{Underspecification} in generic statements in relation to human cognitive predispositions \citep{Cimpian2010GenericSR}. They show that generic statements are inconsistently interpreted, suggesting a potential for discourse manipulation.
We found many instances of generic underspecified claims in \datasetname{}, such as the last example in Table~\ref{tab:pragmaticexamples}. The claim is false \textit{in Britain}, but ill-defined elsewhere, leading to disagreement on the veracity label for the generic statement. This inference type is the reverse of ``\textit{Accommodating Minimally Added Content}'' in hypotheses in \citet{jiang-tacl2022-investigating}, as the claim (the counterpart to the hypothesis in NLI) in our case is less specific than the evidence. 
NLI data is usually collected by hypotheses being written for given premises \citep{williams-etal-2018-broad}, whereas the claims in realistic fact-checking data are generated independently from evidence, which  leads to different inference types being encountered. 

\paragraph{Annotation Scheme}
We employ stratified sampling to select 384 items, ensuring coverage of both rare and frequent veracity annotation combinations.
Each claim is then evaluated with respect to its evidence sentences to determine whether the veracity judgment depends on a specific type of inference or is explicit. 
Initially, a subset of 20 items was double-annotated to assess the consistency of the guidelines, resulting in a Cohen's $\kappa$ agreement of 0.67. 
Subsequently, an additional 364 items were annotated by one of the authors with graduate training in Linguistics. 

\paragraph{Variables and Statistics Measured}

We perform an {\sc ANOVA} variance analysis to examine the relationship between the independent variables (inference types) and the dependent variable (annotator agreement on veracity labels). Interactions are not included due to the non-overlapping nature of the independent variables in the linguistic inference annotation scheme.
Confounders, such as the length of evidence and claim, presence of negation in the claim, and presence of quantifiers in the claim, are added to account for variance unrelated to the hypothesized independent variables. 
These confounders aim to capture aspects of annotator behavior, as increased cognitive load from negation or longer input length might negatively impact annotation quality, while quantifiers could make the claims clearer to the annotators.

\paragraph{Results}

The variance analysis showed an $R^2$ value of 0.367, indicating that a significant portion of the variation in annotator disagreement could be explained by annotators' sensitivity to non-explicitly communicated content in claims or evidence as captured by the independent variables. Table~\ref{tbl:pragmatic-inference-results} presents the significant effects observed in the correlation between the presence of inference types and the level of disagreement. The coefficients in the table reveal that ambiguous content is significantly linked to agreement scores, with the presence of negation in the claim also having a significant effect, likely due to confusion regarding polarity. This corroborates the results of previous work, showing that ambiguity is inherent to linguistic data and therefore annotator disagreement on labels should be incorporated in NLP models.

\begin{table}[htb]
\small
    \centering
    \begin{tabular}{l | c | c}
    Independent variable & coefficient & P-value  \\
    \toprule
    {Implicature} &-0.2989&0.000*\\
    {Presupposition} &-0.3800&0.012*\\
    {Coreference} &-0.2961&0.009*\\
    {Vagueness} &-0.6469&0.000*\\
    {Underspecification} &-0.6111&0.000*\\
    {Probabilistic reasoning} &-0.5590&0.000*\\
    \midrule
    Evidence length  &-0.0356&0.966\\
    Claim length  &1.2642&0.600\\
    Negation in claim  &-0.2209&0.001*\\
    Quantifier in claim  &0.1255&0.119\\
      \bottomrule
    \end{tabular}
    \caption{Results of {\sc ANOVA} showing linguistic inference effects on Krippendorff's $\alpha$ in \datasetname{}. The significant effects are marked with an asterisk.}
    \label{tbl:pragmatic-inference-results}
\end{table}

\section{Experiments}
\label{sec:experiments}

\subsection{Evidence Selection (Ev.)}

The system is tasked with identifying the sentences in a given Wikipedia passage $P=[s_1, ..., s_n]$ that serve as evidence $s_i \in E \subseteq P$ for a claim $c$. 
We use the F1-score over claim-sentence pairs ($c$, $s_i$) for evaluation.
\label{subsec:experiments:evidence-extraction}
We compare four evidence selection methods using binary/ternary evidence annotations with hard or soft labels. Evaluation is performed on \datacertain{} and \datasetname{} datasets.

\paragraph{Models}
We experiment with four evidence selection approaches.
First, following \citet{thorne-etal-2018-fever} and \citet{wadden-etal-2020-fact} we model (Ev.) as a binary classification task. 
Second, we train a model to predict the ternary label for each ($c$, $s_i$). For training, the majority of ``supporting'' and ``refuting'' annotations determines the ternary label, with the overall majority (``supporting'') as tiebreaker, and "neutral" assigned if only neutral annotations exist. Ternary predictions are mapped to binary evidence labels for evaluation. We refer to these evidence selection models as \emph{binary} or \emph{ternary} respectively.
To handle 
the different perspectives by the annotators,
one intuitive approach is to mimic the annotation distribution using distillation \citep{hinton2015distilling,fornaciari-etal-2021-beyond}. Annotation distillation is achieved by minimizing the soft cross-entropy loss between human and predicted distributions. Previous studies directly modeled human annotation probabilities for each class \citep{peterson2019human}, or applied softmax over annotation counts \citep{uma2020case}.
We calculate human probabilities by dividing the frequency of annotations per class by the total number of annotations per instance, as this method proved most effective for \datasetname{} in our initial experiments. We refer to models that distill these probabilities as \emph{distill} models.
A sentence is classified as evidence if the sum of predicted probabilities for ``supporting'' and ``refuting'' 
exceeds a threshold 
chosen by maximizing the evidence F1-score on the dev set, with values ranging from 0 to 0.3 in intervals of 0.01. 
Lastly, we experiment with a regression approach for evidence selection.
We calculate the estimated probability $p_i$ for a sentence $s_i$ being part of the evidence set $E$ based on the ratio of annotators who assigned a non-neutral label.
We train a regression model (denoted as \textit{regr}) to predict the probabilities $p_i$ by minimizing the MSE loss. 

\subsection{Veracity Prediction (Ver.)}
We experiment with soft labels on the entire \datasetname{} and with aggregated labels only on \datacertain{}. Previous studies in fact-checking have used two model architectures. 
The \emph{Pipeline} approach predicts the claim's veracity solely based on selected evidence, as seen in approaches for FEVER.
Following \citet{wadden-etal-2020-fact}, we randomly sample one to two sentences during training only when no evidence sentence exists. During inference, if no evidence is selected, the prediction defaults to neutral.
The second architecture is the \emph{Full-text} approach, where veracity is directly predicted based on the entire evidence document(s) as by \citet{augenstein-etal-2019-multifc} or \citet{park2021faviq}.

\label{subsec:experiments:aggregated}
Fact-checking tasks typically assume single veracity labels
\citep{thorne-etal-2018-fever, schuster-etal-2021-get, park2021faviq}. However, aggregated labels cannot capture the ambiguity in 
\datasetname{}. 
Therefore, our evaluation based on aggregated labels is only applied on \datacertain{}, 
which exhibits higher annotator agreement for the veracity label. We experiment with soft labels using the entire dataset \datasetname{} = \datacertain{} $\cup$ \datauncertain{}.

\subsubsection{Single Label Veracity Prediction}
To aggregate the passage-level veracity annotations 
we employ the Dawid-Skene \citep{dawid1979maximum} method using the implementation of \citet{ustalov2021crowdKit} on \datacertain{}.
Models are assessed based on their accuracy of in predicting the veracity for each ($c$, $P$). Similar to FEVER-score \citep{thorne-etal-2018-fever}, we require models to correctly predict the evidence and veracity label (Ev.+Ver.). We score models via the averaged instance-level product of the evidence F1-score with the accuracy of the veracity label. This results in scores of zero when either the veracity or evidence is incorrect, thereby penalizing the model if it doesn't perform well in both tasks.

\paragraph{Models}
We compare pipeline and full-text models for single-label veracity prediction (\textsc{Single}). We also evaluate a self-correcting version of the pipeline (\textsc{CSingle}), which removes selected evidence if it predicts ``neutral'' as veracity.
Baseline models utilize selected sentences from the ternary evidence selection approach:
The \textsc{Max} baseline selects the stance with the highest probability, while the \textsc{Maj} baseline uses majority voting. 
Sentences are only considered if the predicted probability for a non-neutral label reaches a threshold $t=0.95$.
We determine the threshold $t$  by optimizing the accuracy on the dev set over values ranging from 0 to 1 at intervals of 0.05.

\subsubsection{Soft Labels Veracity Prediction}
\label{sec:experiments:veracity-prediction-soft-labels}
Incorporating diverse annotations in model evaluation is still an open challenge \citep{plank-2022-emnlp}. We use four metrics adapted from recent literature \citep{baan-etal-2022-stop,jiang-tacl2022-investigating}, to score models:
The Human Entropy Calibration Error (\emph{EntCE}) assesses the difference in indecisiveness between humans and model predictions by comparing their distribution entropies at the instance level.
The Human Ranking Calibration Score (\emph{RankCS}) evaluates the consistency of label rankings between predicted and human probabilities at the instance level. We modify RankCS introduced by \citet{baan-etal-2022-stop} to handle multiple valid rankings for veracity labels identically.
The Human Distribution Calibration Score (\emph{DistCS}) is derived from \citet{baan-etal-2022-stop} and quantifies the total variance distance (TVD) between the predicted distribution $\hat{y}$ and the human label distribution $y$. It is calculated as $\textrm{DistCS} = 1 - \textrm{TVD}(\hat{y}, y)$ at the instance level and is the strictest of our metrics.

Our annotations may not fully capture the true human distribution. Hence, we treat veracity prediction as a multi-label classification task. Following \citet{jiang-tacl2022-investigating}, we require models to predict all veracity labels chosen by at least 20\% of the annotators. We evaluate models using the sample-averaged F1-score (\emph{F1}).
For the joint evaluation (Ev.+Ver.), we calculate the point-wise product of the evidence F1-score with the sample-averaged F1-score (w-F1) and DistCS (w-DistCS).

\paragraph{Models}
We examine four models that incorporate different annotations to different extents. The first model, referred to as \textsc{Single} (from §\ref{subsec:experiments:aggregated}), assumes a single veracity label for each ($c$, $P$) instance. 
Additionally, similar to §\ref{subsec:experiments:evidence-extraction} we train annotation distillation models (denoted as \textsc{Distill}) to learn the human annotation distribution.
When no evidence is selected for the pipeline, the prediction defaults to 100\% neutral.
Third, we apply temperature scaling \citep{guo2017calibration} as a method to recalibrate models by dividing the logits by a temperature parameter $t$ before the softmax operation. This technique has demonstrated effectiveness in various NLP tasks \citep{desai-durrett-2020-calibration}. 
We choose $t$ based on the highest DistCS score on the dev set for the trained \textsc{Single} models.
This calibrated model is denoted as \textsc{Temp. Scaling}.
In the case of the pipeline model, if no evidence is selected, the predicted distribution defaults to 100\% neutral.
Finally, we explore a multi-label classification approach. 
Following \citet{jiang-tacl2022-investigating}, we estimate the probability of each class by applying the sigmoid function to the model's logits. Classes with a probability of $p \geq 0.5$ are considered as predicted. When necessary for computing metrics, we generate probability distributions by replacing the sigmoid function with softmax during inference.
We use evidence selection models with ternary labels and annotation distillation as baselines. 
We combine the predicted probabilities of the labels "supporting" ($S$) and "refuting" ($R$) by summing them, resulting in $p^{S+R}=1-p^N$, where $p^N$ represents the predicted probability for "neutral".
We use the predictions based on the sentence with the highest $p^{S+R}$ as the veracity prediction and refer to this baseline as \textsc{MaxEvid}. We only consider sentences with $p^{S+R}\geq t$, where the threshold $t$ is optimized for DistCS on the development set.

\subsection{Implementation}
We employ DeBERTaV3$_\text{large}$ \citep{he2021debertav3} from the Transformers library \citep{wolf-etal-2020-transformers} for both (Ev.\ and Ver.) tasks, including Pipeline and full-text variants.
DeBERTaV3$_\text{large}$ has achieved exceptional performance on the SuperGlue benchmark, including MNLI~\citep{williams-etal-2018-broad} and RTE~\citep{dagan2006pascal}, related to fact-checking.
We use fixed hyperparameters (\texttt{6e-6} learning rate, batch size of 8)\footnote{As proposed for MNLI: \url{https://huggingface.co/microsoft/deberta-v3-large}.} and train for 5 epochs, selecting the best models based on evidence F1-score (Ev.\ classification), MSE (Ev.\ regression), accuracy (Ver.\ single-label), micro F1-score (Ver.\ multi-label), and negative cross-entropy loss (distillation).
DeBERTaV3$_\text{large}$ accommodates both short text snippets and longer sequences, enabling fair comparisons  between all variants. 
In initial experiments, we observed that including the Wikipedia entity and section title enhances performance. We input all to the model via \texttt{[CLS] claim [SEP] evidence @ entity @ title [SEP]} and feed \textsc{[CLS]} embeddings to linear layer for predictions.

\section{Results}

\begin{table}[ht]
\centering
\small
    \centering
    \begin{tabular}{l c | c  c }
    \toprule
    \multicolumn{2}{c|}{\textit{Training}} & \multicolumn{2}{c}{\textit{Evidence F1}} \\
    \textbf{Data} & \textbf{Model} & \textbf{\datasetname{}} & \textbf{\datacertain{}} \\


        \toprule

         & binary & 64.1 \tiny{$\pm 0.2$} & 64.4 \tiny{$\pm 1.2$}\\
        \multirow{2}{*}{\datasetname{}} & ternary & 63.5 \tiny{$\pm 0.7$} & 64.4 \tiny{$\pm 1.3$}\\
        & regr &  64.5 \tiny{$\pm 0.4$} & 63.1 \tiny{$\pm 0.8$} \\
        & distill &  \textbf{65.3} \tiny{$\pm 0.3$} & 63.0 \tiny{$\pm 1.5$} \\

         \midrule
         & binary & 56.4 \tiny{$\pm 1.2$} & 66.2 \tiny{$\pm 0.6$}  \\
        \multirow{2}{*}{\datacertain{}} & ternary & 54.0 \tiny{$\pm 1.9$} & 65.6 \tiny{$\pm 0.5$}  \\
        & regr &  58.2 \tiny{$\pm 2.3$} & \textbf{66.9} \tiny{$\pm 0.4$} \\
        & distill &  57.9 \tiny{$\pm 2.0$} & \textbf{66.8} \tiny{$\pm 0.6$} \\

       \bottomrule
    \end{tabular}
    \caption{Evidence F1-score averaged with standard deviation over five runs.}
    \label{tbl:evidence-evaluation-t1-main}
\end{table}{}

\paragraph{Evidence Selection}
The results in Table~\ref{tbl:evidence-evaluation-t1-main} show that predicting ternary labels provides no advantage over binary evidence labels. This holds true for both training on the entire \datasetname{} and the \datacertain{} subset. However, integrating annotators' uncertainty in evidence selection consistently improves the overall scores.
Training solely on \datacertain{} leads to lower F1-score on the entire \datasetname{}. A possible reason is the different distribution of evidence sentences: In \datacertain{}, evidence sentences constitute only 8.9\% of all sentences. 
These sentences contain on average 52.2\% non-neutral annotations.
In \datasetname{}, evidence is found in 19.9\% of all sentences.
These sentences contain on average 38.8\% non-neutral annotations.
The \emph{distill} approach trained on \datacertain{} performs well in detecting evidence in \datacertain{} (recall=68.8\%, precision=65.2\%), but struggles on \datasetname{} as it fails to detect many evidence sentences on instances from \datauncertain{} (recall=42.8\%, precision=76.5\%). Training on all of \datasetname{} improves the recall of selected evidence, reaching a recall of 80.4\% / 64.3\% and precision of 51.9\% / 68.3\% on \datacertain{} and \datauncertain{}.

\paragraph{Single Veracity Labels}

\begin{table}
    \centering
    \small
    \begin{tabular}{lcc | c c }
    \toprule
   \multicolumn{3}{c|}{\textit{Model}} & \textit{Ver.} & \textit{Ev.+Ver.} \\
    \midrule
    \multicolumn{1}{c}{\textbf{Train}}&\textbf{Ev.} & \textbf{Ver.} & \textbf{Acc.} & \textbf{w-Acc.} \\
    \midrule
    \multicolumn{1}{c}{--}&\textit{oracle} & \textsc{Maj.} & \textit{98.5} & \textit{98.5} \\
    \multicolumn{1}{c}{--}&\textit{oracle} & \textsc{Single} & 97.1  &  97.1\\
    \midrule
    &ternary & \textsc{Maj.} & 91.4 & 85.2 \\
    \multirow{2}{*}{\datacertain{}}&ternary & \textsc{Max.} & 91.5 & 85.2  \\
    &regr & \textsc{Single} & 94.0  & 83.3  \\
    &regr & \textsc{CSingle} & 94.0  & \textbf{88.2}  \\
    &-- & \textsc{Single} & \textbf{94.1}  & --  \\
    \midrule
    &ternary & \textsc{Maj.} & 89.0 & 82.8  \\
    \multirow{2}{*}{\datasetname{}}& ternary & \textsc{Max.} & 89.1 & 82.9   \\
    &binary & \textsc{Single} & 94.1  & 77.0  \\
    &binary & \textsc{CSingle} & 94.1  & \textbf{88.0}  \\
    &-- & \textsc{Single} & \textbf{94.4}  & --  \\

    \bottomrule
    \end{tabular}
    \caption{Averaged veracity prediction results on \datacertain{} over aggregated single labels over five runs.}
    \label{tbl:main-results-aggregated-label}
\end{table}

We evaluate single label classification models for veracity prediction, selecting the best evidence selection methods from Table~\ref{tbl:main-results-aggregated-label}. The \textsc{Maj} and \textsc{Single} models achieve high accuracy when provided with oracle evidence.
When using automatically selected evidence, \textsc{Single} outperforms the baselines on (Ver.) but performs worse on the joint score (Ev.+Ver.):
One possible explanation is that our baselines cannot predict ``neutral'' when evidence sentences are selected. This is beneficial on \datacertain{} where 96.6\% of all instances with evidence sentences have non-neutral veracity labels. The trained \textsc{Single} model, however, can incorrectly predict ``neutral'' even when evidence is correctly identified.
For comparison, assuming single labels on \datauncertain{}, 37.2\% of instances have a neutral veracity along with supporting or refuting evidence sentences. Training on \datasetname{} improves performance on aggregated labels in \datacertain{}, especially for the full-text model that avoids errors from evidence selection.

\begin{figure}
\small
    \centering
    \includegraphics[width=\linewidth]{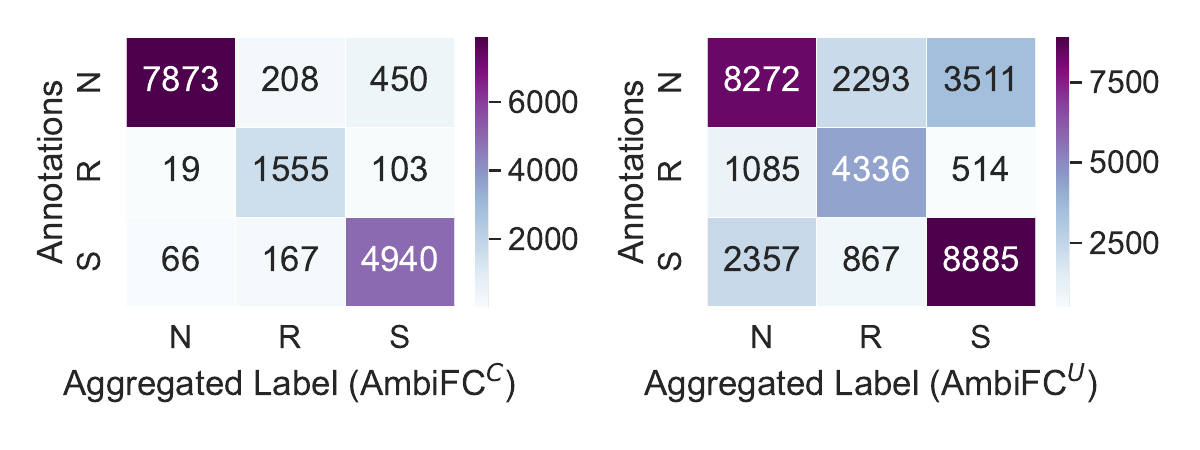}
    \caption{Annotations that are (not) considered by aggregated labels on the respective test sets.}
    \label{fig:aggregated-labels-and-annotations}
\end{figure}
High scores on aggregated labels may not comprehensively represent all valid perspectives \citep{prabhakaran-etal-2021-releasing, fleisig2023majority}. In the test set, 6.6\% of annotations in \datacertain{} are ignored by the aggregated labels (Figure~\ref{fig:aggregated-labels-and-annotations}; left). 
The single-label prediction of the full-text model trained on \datacertain{} aligns with 87.3\% of the veracity annotations.
In comparison, aggregated veracity labels in \datauncertain{} would  capture only 66.9\% of all annotations (Figure~\ref{fig:aggregated-labels-and-annotations}; right).
The \datasetname{}-trained full-text model only agrees with 57.1\% of them when predicting single labels (with a computed accuracy of 68.8\%).
Both highlight the importance of annotation-based evaluations throughout \datasetname{}.

\paragraph{Soft Veracity Labels}
\begin{table*}[ht]
\centering
\small
    \centering
    \begin{tabular}{cc | c c c c| c c}
    \toprule
    \multicolumn{2}{c|}{\textit{Model}} & \multicolumn{4}{c|}{\textit{Ver.}} & \multicolumn{2}{c}{\textit{Ev. + Ver.}} \\
    \textbf{Ev.} & \textbf{Ver.} & \textbf{EntCE}$\downarrow$ & \textbf{RankCS}$\uparrow$& \textbf{DistCS}$\uparrow$ & \textbf{F1}$\uparrow$ & \textbf{w-DistCS}$\uparrow$ & \textbf{w-F1}$\uparrow$ \\ 
    \midrule

    \multicolumn{2}{c|}{\textit{avg. distribution}} & .568 & .529 & .597 & .747 & .215 & .267\\
    \midrule
    
    ternary & \textsc{MaxEvid.} & .305 \tiny{$\pm 0.03$} & .644 \tiny{$\pm 0.01$} & .701 \tiny{$\pm 0.03$} & .730 \tiny{$\pm 0.03$} & .546 \tiny{$\pm 0.02$} & .506 \tiny{$\pm 0.03$} \\
    distill & \textsc{MaxEvid.} & .223 \tiny{$\pm 0.01$} & .712 \tiny{$\pm 0.01$} & .793 \tiny{$\pm 0.00$} & .850 \tiny{$\pm 0.00$} & .574 \tiny{$\pm 0.01$} & .593 \tiny{$\pm 0.00$} \\
    \midrule


    \textit{oracle} & \textsc{Single} & .289 \tiny{$\pm 0.03$} & .779 \tiny{$\pm 0.01$} & .787 \tiny{$\pm 0.01$} & .800 \tiny{$\pm 0.01$} & .787 \tiny{$\pm 0.01$} & .800 \tiny{$\pm 0.01$} \\
    \textit{oracle} & \textsc{Temp. Scaling} & .175 \tiny{$\pm 0.00$} & .779 \tiny{$\pm 0.01$} & .840 \tiny{$\pm 0.00$} & .842 \tiny{$\pm 0.01$} & .840 \tiny{$\pm 0.00$} & .842 \tiny{$\pm 0.00$} \\
    \textit{oracle} & \textsc{Multi} & .244 \tiny{$\pm 0.01$} & .792 \tiny{$\pm 0.00$} & .810 \tiny{$\pm 0.00$} & \textbf{.915} \tiny{$\pm 0.00$} & .810 \tiny{$\pm 0.00$} & \textbf{.915} \tiny{$\pm 0.00$} \\
    \textit{oracle} & \textsc{Distill} & \textbf{.146} \tiny{$\pm 0.00$} & \textbf{.801} \tiny{$\pm 0.00$} & \textbf{.867} \tiny{$\pm 0.00$} & .891 \tiny{$\pm 0.00$} & \textbf{.867} \tiny{$\pm 0.00$} & .891 \tiny{$\pm 0.00$} \\
    \midrule


    distill & \textsc{Single} & .306 \tiny{$\pm 0.02$} & .744 \tiny{$\pm 0.01$} & .760 \tiny{$\pm 0.01$} & .777 \tiny{$\pm 0.01$} & .552 \tiny{$\pm 0.01$} & .543 \tiny{$\pm 0.00$} \\
    distill & \textsc{Temp. Scaling} & .244 \tiny{$\pm 0.01$} & .744 \tiny{$\pm 0.01$} & .795 \tiny{$\pm 0.00$} & .812 \tiny{$\pm 0.01$} & .584 \tiny{$\pm 0.00$} & .567 \tiny{$\pm 0.01$} \\
    distill & \textsc{Multi} & .270 \tiny{$\pm 0.01$} & .755 \tiny{$\pm 0.01$} & .782 \tiny{$\pm 0.01$} & .881 \tiny{$\pm 0.01$} & .566 \tiny{$\pm 0.01$} & \textbf{.615} \tiny{$\pm 0.01$} \\
    distill & \textsc{Distill} & \textbf{.214} \tiny{$\pm 0.00$} & .764 \tiny{$\pm 0.00$} & \textbf{.826} \tiny{$\pm 0.00$} & .862 \tiny{$\pm 0.00$} & \textbf{.603} \tiny{$\pm 0.01$} & .601 \tiny{$\pm 0.00$} \\

    \midrule

    %
    %
    %
    %
    -- & \textsc{Single} & .302 \tiny{$\pm 0.02$} & .755 \tiny{$\pm 0.00$} & .765 \tiny{$\pm 0.01$} & .782 \tiny{$\pm 0.01$} &  -- & -- \\
    -- & \textsc{Temp. Scaling} & .264 \tiny{$\pm 0.00$} & .755 \tiny{$\pm 0.00$} & .783 \tiny{$\pm 0.01$} & .801 \tiny{$\pm 0.01$} &  -- & -- \\
    -- & \textsc{Multi} & .249 \tiny{$\pm 0.01$} & .764 \tiny{$\pm 0.00$} & .795 \tiny{$\pm 0.00$} & \textbf{.884} \tiny{$\pm 0.00$} & -- & -- \\
    -- & \textsc{Distill} & .228 \tiny{$\pm 0.00$} & \textbf{.773} \tiny{$\pm 0.01$} & \textbf{.826} \tiny{$\pm 0.00$} & .867 \tiny{$\pm 0.00$} & -- & -- \\

       \bottomrule
    \end{tabular}
    \caption{Results on \datasetname{} averaged over five runs.  All models are trained on \datasetname{}. 
    }
    \label{tbl:main-results-soft-label}
\end{table*}{}

We report the results on \datasetname{} in Table~\ref{tbl:main-results-soft-label}. 
While \textsc{Single} models are not optimized for metrics over soft labels, they serve as informative baselines. Applying temperature scaling 
significantly boosts performance on most metrics, particularly EntCE. 
\textsc{Multi} and \textsc{Distill} outperform other models on various metrics, with each excelling in metrics aligned with their respective optimization objectives. 
The pipeline approach is comparable to the full-text approach in terms of DistCS, while also providing a rationale for predictions and room for improvement through better evidence selection methods (as indicated by \emph{oracle} evidence).  The sentence-level baselines of annotation distillation perform  well, but cannot compete with models trained for veracity prediction.

\begin{table}[ht]
\centering
\small
    \centering
    \begin{tabular}{l|cc}
    \toprule
    & \multicolumn{2}{c}{Trained} \\
    Evaluated & \textbf{\datacertain{}} & \textbf{\datasetname{}} \\
    \midrule
    \textbf{\datacertain{}} & \textbf{.928} & .905\\
    \textbf{\datacertain{}} (5+) & .804 & \textbf{.824}\\
    \textbf{\datauncertain{}} & .642 & \textbf{.751}\\
    \textbf{\datasetname{}} &.781 & \textbf{.826}\\

       \bottomrule
    \end{tabular}
    \caption{\textit{DistCS}$\uparrow$ evaluated across different subsets. \textit{\datacertain{} (5+)} refers to all instances of \datacertain{} with at least five annotations.}
    \label{tbl:soft-labels-subset}
\end{table}{}
The performance of the top-performing pipelines (based on DistCS) is examined on different subsets in Table~\ref{tbl:soft-labels-subset}.  Additionally training on ambiguous instances from \datauncertain{} improves performance across all subsets, except for \datacertain{}. This discrepancy may be attributed to the abundance of fully neutral instances within \datacertain{} -- which do not exist in \datauncertain{}. Performance on instances with 5+ annotations benefits from the inclusion of ambiguous instances. The notable performance gap between \datasetname{} and the ambiguous claims in \datauncertain{} underscores the challenge posed by these ambiguous cases.

\section{Analysis}

\paragraph{Errors by Linguistic Category}



\begin{figure}
\small
    \centering
    \includegraphics[width=\linewidth]{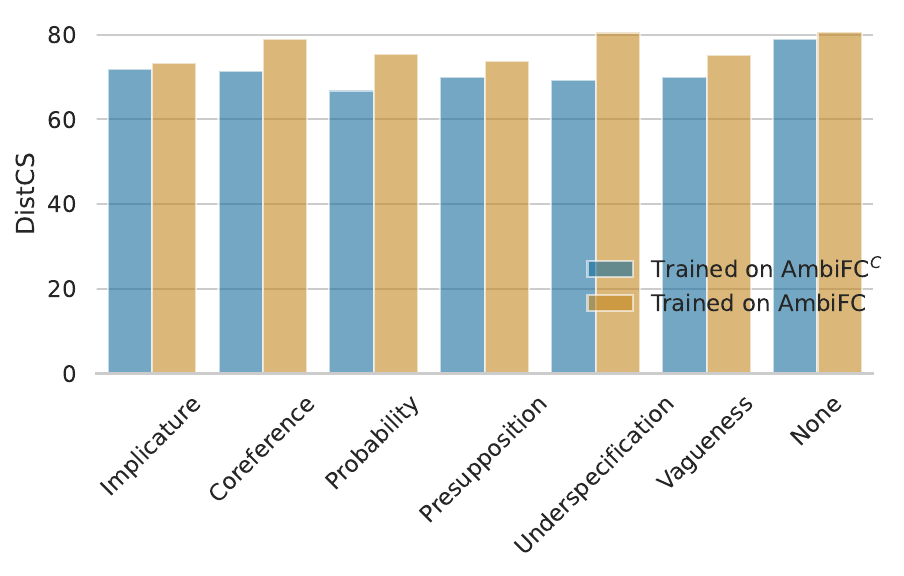}
    \caption{Performance of the Annotation Distillation model on different linguistic categories, separated by the training data used: \datacertain{} and \datasetname{}.}
    \label{fig:error_analysis}
\end{figure}

Model performance varies depending on which lexical, pragmatic and discourse inference types are present in the items. 
We compare the predictions of the best performing model (Annotation Distillation, last row in Table~\ref{tbl:main-results-soft-label}) trained on \datacertain{} and \datasetname{}, and separate the results per linguistic category (Figure~\ref{fig:error_analysis}).
The results corroborate the analysis in §\ref{sec:analaysis:disagreement-analysis}, as the smallest difference between the models trained on \datacertain{} and \datasetname{} is seen with items without linguistic cues for ambiguity.
Furthermore, the largest difference appears in the subsets of the development set which contain Underspecification, Vagueness, Probabilistic Enrichment and Coreference, and 
the first three of these categories have the strongest correlation with annotator disagreement, as seen in Table~\ref{tbl:pragmatic-inference-results}. This suggests that the model performs better on the more ambiguous items when it has seen such items in training.
Furthermore, Underspecification, Vagueness and Coreference have a lower agreement in the \datacertain{} subset as compared to the overall agreement in the \datasetname{}. This suggests that the annotators are often not aware of the presence of alternative interpretations in these classes, which could also be the reason for these items being more difficult for the model to learn.  

\paragraph{Correct Probabilities by Veracity Labels}
\begin{figure}
\small
    \centering
    \includegraphics[width=\linewidth]{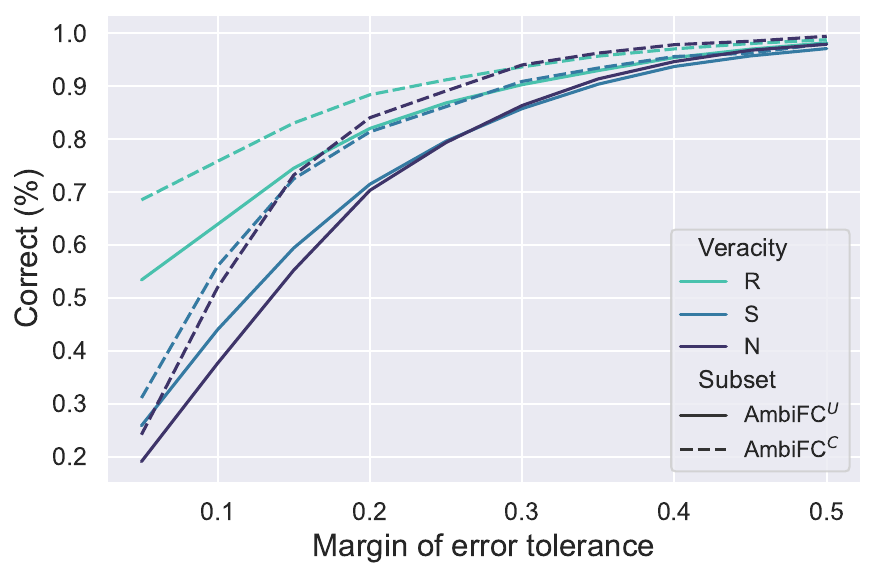}
    \caption{Correct Veracity Estimation by allowing Errors within the margin of the threshold.}
    \label{fig:label-probability-by-t}
\end{figure}
We analyze how accurately the \textsc{Distill} pipelines trained on \datasetname{} predict veracity label probabilities in Figure~\ref{fig:label-probability-by-t}. Predictions are considered correct if the difference between human and predicted probabilities falls within the tolerance $t$ on the x-axis. 
With a tolerance of $t=0.15$, the pipeline accurately predicts the probability for 70\% of instances across all labels in \datacertain{}. However, the performance is consistently lower
in \datauncertain{}, highlighting the greater challenge posed by this subset.
The model performs best in predicting the probability for ``refuting'' labels on both subsets. 
This is likely because it assigns a lower probability to this less common label. 
When no refuting annotations exist, the average error is $0.04$. However, when refuting annotations are present, the error increases to $0.19$.

\paragraph{Contradictory Evidence Interpretations}

\begin{figure}
\small
    \centering
    \includegraphics[width=\linewidth]{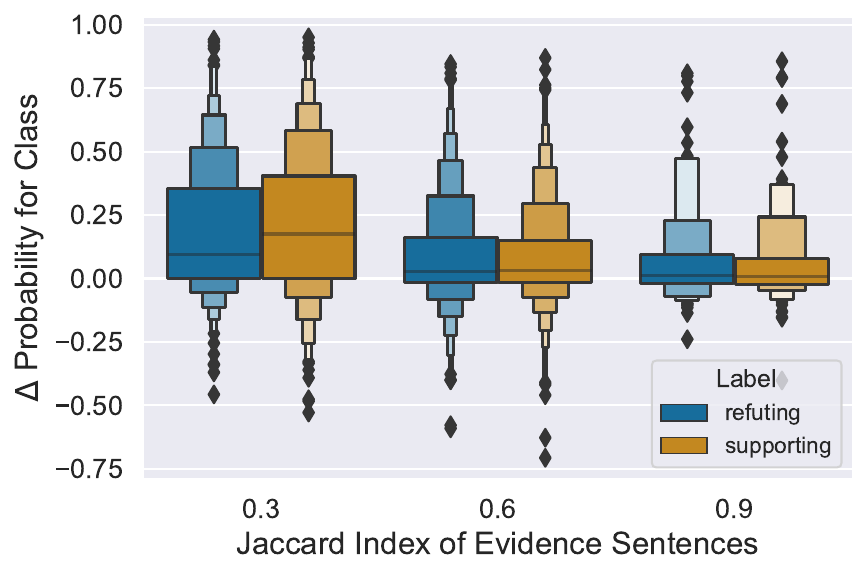}
    \caption{Probability differences of the correctly predicted class when only providing evidence for one of both veracity labels (\emph{S} or \emph{R}).}
    \label{fig:different-evidence-provided}
\end{figure}
Following our observations in §\ref{subsec:analysis:different-evidence} we analyze whether models learn the subtle differences between different evidence sentences for different veracity interpretations. 
We analyze the predictions of a \textsc{Distill} pipeline model ($\mathcal{M}$) by inputting evidence sentences annotated with supporting ($E^S$) \emph{or} refuting ($E^R$) veracity labels separately. 
A model that captures the subtle differences would assign high probabilities to the refuting veracity label $R$ given $E^R$, and low probabilities to $R$ given $E^S$. 
We input the claim $c$ and evidence $E$ into $\mathcal{M}$ to predict the probability $p^R$ for the veracity label $R$ via $p^R=\mathcal{M}(c, E)$.
We measure 
the different effect of $E^R$ and $E^S$ for both veracity labels $R$ and $S$
as $\Delta p^R = \mathcal{M}(c, E^R) - \mathcal{M}(c, E^S)$ and $\Delta p^S = \mathcal{M}(c, E^S) - \mathcal{M}(c, E^R)$.
In Figure~\ref{fig:different-evidence-provided}, we examine all 1,352 test instances from \datasetname{} with both supporting and refuting veracity annotations. To address cases where similar sentences are selected for $E^R$ and $E^S$, we group samples based on their similarity using the Jaccard Index.
Presenting only $E^R$ or $E^S$ generally increases the probability of the correct class. On average, the $\Delta p$ score is at 11.9\%, and decreases with 
more overlap
between sentences in $E^R$ and $E^S$.

\section{Conclusions}
We present \datasetname{}, a fact-checking dataset with annotations for evidence-based fact-checking, addressing the inherent ambiguity in real-world scenarios.
We find that annotator disagreement signals ambiguity rather than noise 
and provide explanations for this phenomenon through an analysis of linguistic phenomena. 
We establish baselines for fact-checking ambiguous claims, leaving room for improvement, particularly in the area of evidence selection. By publishing \datasetname{} along with its annotations, we aim to contribute to research integrating annotations into trained models.

\paragraph{Limitations}
Claims in \datasetname{} are based on real-world information needs. They are not collected from real-world sources and differ from claims seen as check-worthy by human fact-checkers.
\datasetname{} lacks evidence retrieval beyond the passage level. It contains different veracity labels for the same claim given different passages, without overall verdict. Models trained on \datasetname{} are constrained to this domain and only address partial aspects of complete fact-checking applications, as defined by \citet{guo-etal-2022-survey}.

\iftaclpubformat

\section*{Acknowledgments}
This work was supported through a gift from Google as well as the donation of cloud compute credits. The authors wish to thank Dipanjan Das for his advice and support. 
The authors would like to thank the anonymous reviewers and the Action Editor for their valuable feedback and discussions.
The authors would like to thank Jan Buchmann, Sukannya Purkayastha and Jing Yang for their valuable feedback on an early version of this publication.
Conditional scoring with F1 arose from an idea during a conversation with Jonty Page. 
Max Glockner is supported by the German Federal Ministry of Education and Research and the Hessian Ministry of Higher Education, Research, Science and the Arts within their joint support of the National Research Center for Applied Cybersecurity ATHENE. 
Ieva Staliūnaitė is supported by Huawei.
James Thorne is supported by Institute of Information \& communications Technology Planning \& Evaluation (IITP) grant funded by the Korea government (MSIT) (No.2019-0-00075, Artificial Intelligence Graduate School Program (KAIST)).
Gisela Vallejo is supported by the graduate research scholarship from the Faculty of Engineering and Information Technology, University of Melbourne.
Andreas Vlachos is supported by the ERC grant AVeriTeC (GA 865958) and the EU H2020 grant MONITIO (GA 965576).
\else
\fi

\bibliography{references}
\bibliographystyle{acl_natbib}


\onecolumn

\appendix


\end{document}